\documentclass[letterpaper]{article} 
\usepackage{aaai22}  
\usepackage{times}  
\usepackage{helvet}  
\usepackage{courier}  
\usepackage[hyphens]{url}  
\usepackage[dvipdfmx]{graphicx}
\usepackage{natbib}  
\usepackage{caption} 
\usepackage{algorithm}
\usepackage{algorithmic}

\usepackage{newfloat}
\usepackage{listings}
\DeclareCaptionStyle{ruled}{labelfont=normalfont,labelsep=colon,strut=off} 
\floatstyle{ruled}
\newfloat{listing}{tb}{lst}{}
\floatname{listing}{Listing}

\title{Recognition of All Categories of Entities by AI}
\author{
    Hiroshi Yamakawa \textsuperscript{\rm 1,\rm2}, 
    Yutaka Matsuo \textsuperscript{\rm 1},\\
}
\affiliations{
    \textsuperscript{\rm 1} The University of Tokyo\\

     \textsuperscript{\rm 2} The Whole Brain Architecture Initiative

}

\usepackage{bibentry}

\begin{document}

\maketitle

\begin{abstract}

Human-level AI will have significant impacts on human society. However, estimates for the realization time are debatable. To arrive at human-level AI, artificial general intelligence (AGI), as opposed to AI systems that are specialized for a specific task, was set as a technically meaningful long-term goal. But now, propelled by advances in deep learning, that achievement is getting much closer. Considering the recent technological developments, it would be meaningful to discuss the completion date of human-level AI through the "comprehensive technology map approach," wherein we map human-level capabilities at a reasonable granularity, identify the current range of technology, and discuss the technical challenges in traversing unexplored areas and predict when all of them will be overcome. This paper presents a new argumentative option to view the ontological sextet, which encompasses entities in a way that is consistent with our everyday intuition and scientific practice, as a comprehensive technological map. Because most of the modeling of the world, in terms of how to interpret it, by an intelligent subject is the recognition of distal entities and the prediction of their temporal evolution, being able to handle all distal entities is a reasonable goal. Based on the findings of philosophy and engineering cognitive technology, we predict that in the relatively near future, AI will be able to recognize various entities to the same degree as humans.

\end{abstract}

\section{Introduction} 
For better or worse, human-level AI will have significant impact on human society. Thus, the debate over when it will emerge has gained considerable attention. However, estimates for the realization time range from a few decades to the 22nd century and beyond.
In the 2000s, with the technology at hand, it was not easy to envision a complete human-level AI. 
Based on this perspective, it was estimated that AI would be realized in the 2040s, calculating the point at which the computational resources used for AI would be at the human level\cite{Kurzweil2005-lj} .

The technical problem with AI systems prevalent then was that they were specialized for a specific task and could not solve a wide range of tasks in a manner similar to humans. Therefore, artificial general intelligence (AGI) was set as a technically meaningful long-term goal, where one AI system can solve many tasks \cite{Goertzel2014-oy}.
However, a comprehensive list of the tasks to be solved is necessary to estimate when advanced AI will be realized in a general-purpose manner. Therefore, as a first step toward the realization of AGI, attempts were made to list the task-solving abilities that a human-level AI should acquire during development \cite{Adams2012-xs}. However, the exhaustiveness of the list cannot be guaranteed because there are infinite variations of possible tasks. 

By contrast, the technological landscape of AI had changed dramatically since 2013, when deep learning began to develop.
By the late 2010, machine learning AIs (ML-AIs) could complete most individual tasks for which sufficiently training data were available.
In the 2020s, large-scale ML-AI has demonstrated that a single AI can perform many tasks, and research is underway to use it for knowledge transfer between tasks.  Recent large-scale language models have demonstrated human-level capabilities in a relatively broad range of language processing tasks disconnected from real-world actions and perceptions. 


Considering these technological developments, it would be meaningful to discuss the completion date of human-level AI through the "comprehensive technology map approach" as follows:
First, we develop a map covering human-level capabilities at a reasonable granularity from a certain technological perspective.
Next, on the map, the current range should be identified, and its success factors should be analyzed.
Subsequently, we should discuss the technical challenges in traversing the unexplored areas and predict when all of them will be overcome.

A precedent for this is the discussion of dual-process theory as a comprehensive technology map.
In summary, this argument states that deep learning has already realized System 1 level, which is that it can respond fast, although System 2 level, a deliberation requiring sequential inference, is to be developed in the future.
There are two weak points with this argument. One is that the technology map has only two nodes, System 1 and System 2, which makes the map extremely coarse-grained.
The other is that the characterization of System 1 and System 2 differs significantly from researcher to researcher.

Therefore, this paper proposes the option of using the ontological sextet t\cite{Smith2005-mz} (see below for details), an exhaustive map of existence, as a comprehensive technical map.
The primary capacity for a subject to be intelligent is modeling the world, which primarily relies on recognizing distal entities and predicting their temporal evolution \cite{Suzuki2015-fm, Saigo2019-bv}.
Hence, an AI with the ability to recognize all categories on the ontological sextet is a reasonable goal by which one could judge the attainment of human-level AI.

The subject's ability to handle various entities means that it can process to find various patterns that can be considered preserved in the world. This includes physical reality, property (color, size, etc.), events, and all other entities we perceive.
In this paper, we call the process of perceiving existence "entification.
And the benefits of the subject's ability to entification are great. First, imagining a real task makes it possible to design drawings and use tools. Furthermore, it makes it possible to give instructions and communicate using symbols.

\subsection{Ontological Sextet}
\begin{figure}[t]
\centering
\includegraphics[width=0.99\columnwidth]{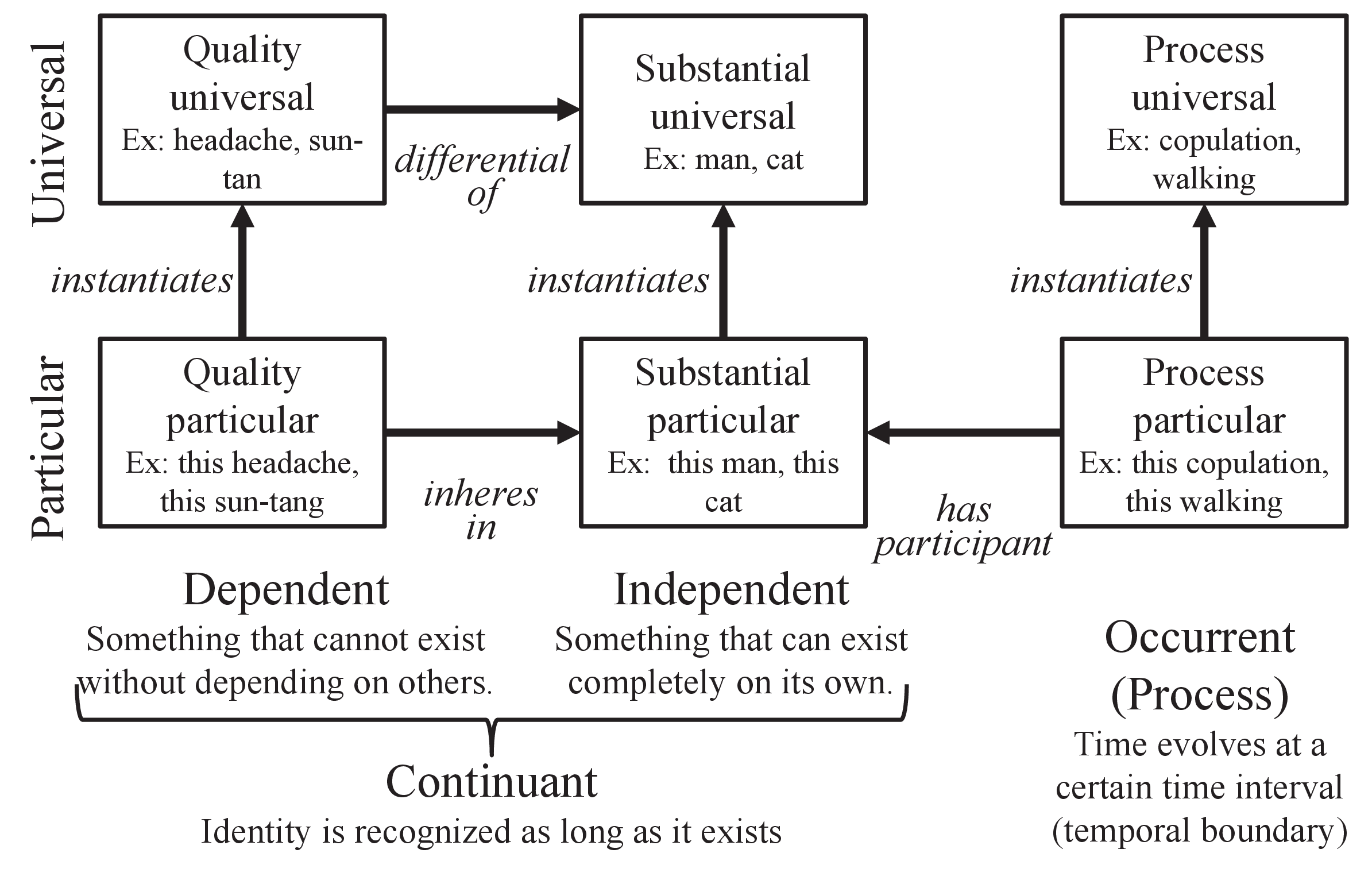} 
\caption{Ontological Sextet: Categories are a combination of substantial / quality / process and particular / universal. }.
\label{fig1}
\end{figure}

Since Aristotle's time, there have been attempts to create mutually disjoint and exhaustive categorization of existents.
We utilizes the ontological sextet \cite{Arp2015-nw}, which encompasses entities in a way that is consistent with our everyday intuition and scientific practice.

The ontological sextet is composed of six categories, which have a two-row, three-column structure described as follows (see fig.\ref{fig1}):
First, they are classified as continuant, which remains identical as long as it exists, and occurrent (process), which develops across a specific time interval.
Continuants are further divided into substantial, which can exist alone, and quality, which exist in dependence on substantial.
For each of these three categories, there is the particular, which exists concretely without stating what it is in a sentence, and the universal, which says what it is in a sentence and is shared by multiple particulars.
In other words, “to be is to be the value of a bound variable” \cite{Quine1948-pq}
In addition, following formal relationships exist between the categories: substantial particulars are participants in process particular;
quality particular is inherent in substantial
 particular; and quality universal differentiates substantial universal.
Furthermore, all universals are formed by instances of corresponding particulars. The aforementioned formal relationship shows that substantial particular is the basis of other categories.
Substantial particular is a concrete object in front of you, such as this person or this cat.
In contrast, substantial universal is a kind, such as a person or a cat.
Quality particular is an entity that inheres in substantial particular, such as this white, this tan, etc.
Process particular is a process in which the substantial particular participates, such as this walk or this collision.

Section 2 summarizes the reasons why entification is needed and the required functions.
In Section 3, the mechanism of entification is described.
In Section 4, current and future development of the entification on ontological sextet is surveyed.
Section 5 provides a summary.

\section{Basis of entification process}

This section clarifies basis of entification for substantial particular, which is the basic part of the ontological sextet.

\subsection{Request to compress the world}

Definitions of intelligence vary widely \cite{Legg2007-dl}; however, in general, intelligence is "the ability to solve complex problems” \cite{Tegmark2017-zo}.
Most of it will rely on "the ability to make predictions about the world" \cite{Conant1970-mh, Friston2006-ry}.

Here, we review the quantitative assumptions necessary to model the world and make predictions.
First, the information that a subject (part of the world) can obtain from sensors and its computational power is finite.
Second, some order exists in the world owing to the uniformity principle.
Third, the entire world is much vaster than the subject's capabilities can handle.
Fourth, compression and prediction can be performed based on informational redundancy.


Considering these assumptions, a subject compresses information of the world, increasing its prediction ability.
The compression process transforms a given set of cases into a more minimalistic representation.
In addition, the goal of intelligence is to increase predictability, and compression is a tool for that. 
Hence, the perspective of Epicurus' multiple explanatory principles, pursuing diversity, is also essential.
\subsection{Sub-functions to construct entification}

Here, we investigate the functions necessary for an entification for the substantial particular and the quality particular.

\subsubsection{ Ability to recognize diachronic identical entities } 

The condition for an existing object to be identical over time ("diachronic identity") has been examined in philosophy.
This answers the question, "What does it mean for an object or substance to remain the same?”
Entification must recognize a set of observed patterns that satisfy diachronic identity as one entity.


Even in contemporary philosophy, no established theory for diachronic identity exists.
We adopt a condition of diachronic identity that seems relatively moderate and appeals to the following notion of continuity\cite{Kurata2017-rm}.


\smallskip
\noindent \underline{ Condition of diachronic identity:}\\
$x$ at point $t$ and $y$ at time $t'$ are diachronically identical and is a conjunction of the following three
\begin{itemize}
\item Spatiotemporal localization: $x$ and $y$ are spatiotemporally continuous.
\item Qualitative continuity: $x$ and $y$ are qualitatively continuous.
\item Kind-specific sameness: $x$ and $y$ belong to the same kind.
\end{itemize}
\smallskip

In short, the condition of diachronic identity is established when some regularity is observed in the physical world, including qualitative changes localized in space--time.

Here, we will discuss the partial functions of entification corresponding to each condition.
The function of extracting a continuous region on the observed data that corresponds to the spatiotemporal region occupied by an individual object from the observed data is called the "single-out function." Entification can satisfy the spatiotemporal localization condition by having this function.

Next, we will consider the essential aspects of existence: the kind-specific sameness and qualitative continuity.
We assume that the single-out and alignment processes have already been applied. 
Regarding kind-specific identity, it should be noted that the essence of an object varies from kind to kind.
For example, the sameness of “dogs" as living things, that of "boats" as artificial objects, and that of "rivers" as natural objects are different.
Therefore, objects belonging to different kinds have different conditions of diachronic identity\cite{Wiggins1967-jm,Yokoro2021-ij}.
To satisfy the condition of diachronic identity, it is necessary to have a kind-classification function that determines which kind the observed object belongs to and a criterion of kind-specific sameness that determines whether the sameness criteria regarding the same kinds are satisfied.
The fulfilment of the condition of qualitative continuity can then be confirmed by the fact that the kind-classification function does not change its kind-classification even if the observed information about an object changes.


\subsubsection{Function to obtain observation data in table format} 

To compress information, the data must be in a table format that enables inductive inference. 
The table format consists of columns(columns for each attribute), cases(rows), and fields (elements within a case). A value is stored in each field. A primary key exists as a column that uniquely identifies each case. The cases and  columns are sets, and the rows and columns are in arbitrary order.
This format corresponds to a table in the relational database.
Therefore, a case set sampling function is always required to obtain tabular data.

Data obtained from remote sensors are often in table format as design specifications. For example, in most video data, the video is 30 frames per second, and the number of columns has the size of the number of pixels  $\times$ the number of channels. In other words, the set of cases is obtained as designed by isochronous sampling.

However, in the case of counting objects by moving one's gaze across an image, as in "Find Wally," counting a single wally corresponds to a single instance.
In the case of spoken language recognition, units such as phonemes, words, sentences, paragraphs, and sentences are considered a single instance.
This function of dynamically selecting cases based on some clues is called the case set sampling function.

\subsubsection{ General properties of remote sensing} 

It is generally assumed that the observed data obtained by most remote sensors satisfy the following conditions to capture some sort of invariant pattern to distant things.

\smallskip
\noindent  \underline{ General  properties of remote sensing}\\
In general, the data acquired by a remote sensor is a table consisting of a set of mutually positioned (specification relation) and comparable columns.
\smallskip

A typical example is a signal of intensity $I(x(t))$ received at the time $t$ at position $X$.
In other words, in the set of values $I$ that the sensor retrieves as a given case (time $t$), they are comparable to each other and positioned at $x$.


\subsubsection{ Alignment function to correct distorted observations of distant things} 

\begin{figure}[t]
\centering
\includegraphics[width=0.9\columnwidth]{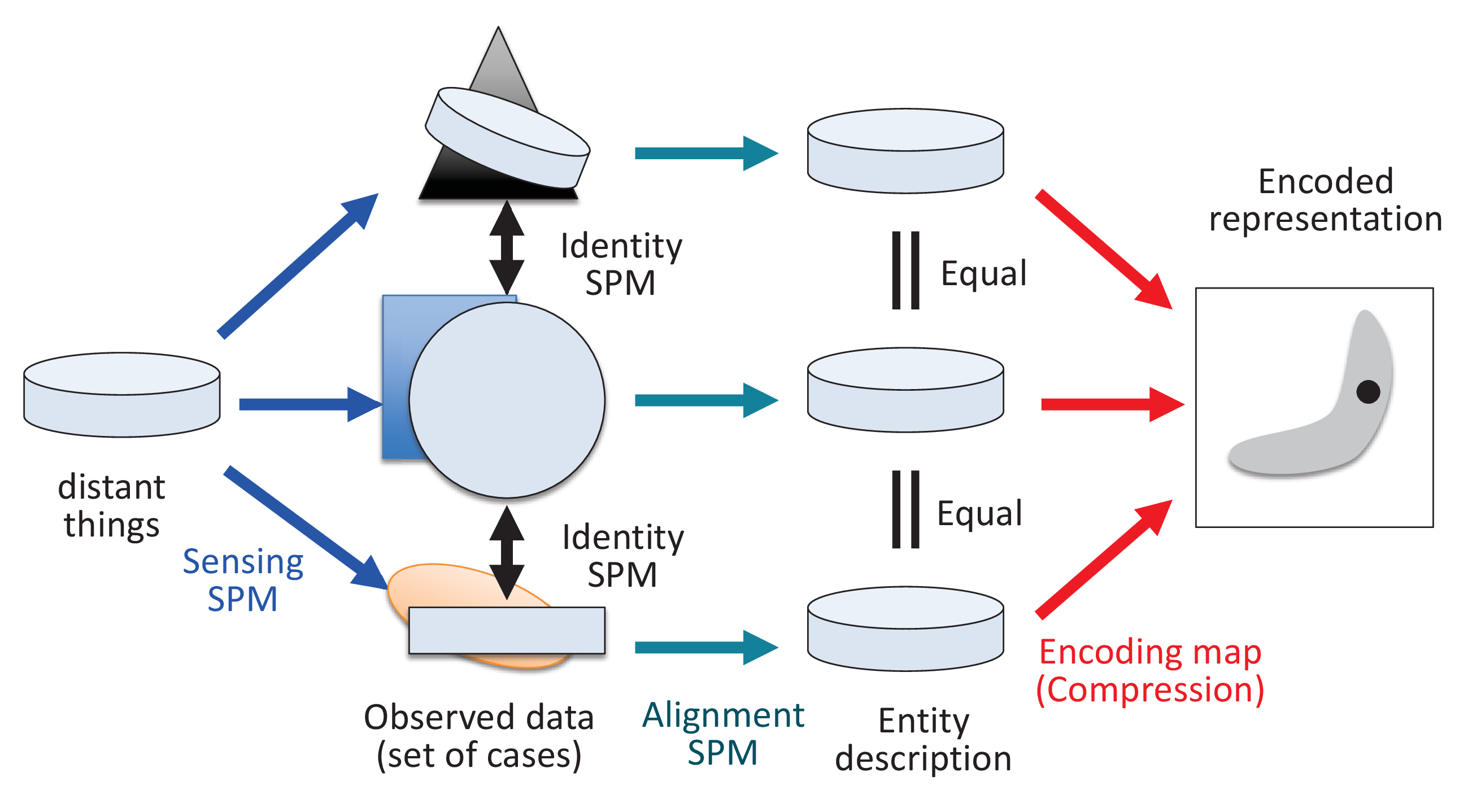} 
\caption{Principle of Entification: Entification occurs through a variety of remote sensors, although it is best illustrated by the visual information example}.
\label{fig2}
\end{figure}

The essence of entification, which recognizes the existence of distal objects, is to identify the patterns preserved in distal things by processing information obtained from remote sensor observations.
The outline of this process is explained as a three-step transformation in Fig. \ref{fig2}: sensing SPM, alignment SPM, and encoding. Here, because the observed data obtained by sensing has the general properties of remote sensing, a structure-preserving map (SPM) related to it can be used.

Sensing SPM refers to mapping patterns obtained from distant things to observed data.
Even if the distant object is invariant, the pattern obtained as observation data on the sampled table changes depending on the relative positional relationship (distance, direction, etc.) with the object to be observed.
For example, in the case of a camera, the sensing SPM is obtained through a mechanism in which reflected light from the object is refracted by a lens and converged on an array sensor.
Partial patterns of observed data in different cases originate from the same distant object and thus form equivalence classes that are matched via the appropriate SPM (Identity SPM in Fig. \ref{fig2})

Alignment SPM aligns some of the patterns in the observation data and transforms them into an alignment table representing the entity description.
This is the alignment function, i.e., the function of extracting and aligning partial patterns preserved in the object from the equivalence class as diversified by the indefiniteness of the observation by the remote sensor.
This is precisely the function that reflects the philosopher Quine's words, "There is no entity without identity" \cite{Quine1969-mr}.
From this positioning, alignment SPM is, in principle, the inverse transformation of sensing SPM.

Encoding maps multiple equal entity descriptions on an aligned table to a specific position in the encoded representation.
This is concerned with the compression requirement and diachronic identity mentioned earlier and is the basis of the kind-classification and kind-specific sameness functions.

\section{Mechanism of Entification}

Here, the entification mechanism for realizing functions discussed in the previous section is proposed.

\begin{figure*}[t]
\centering
\includegraphics[width=0.9\textwidth]{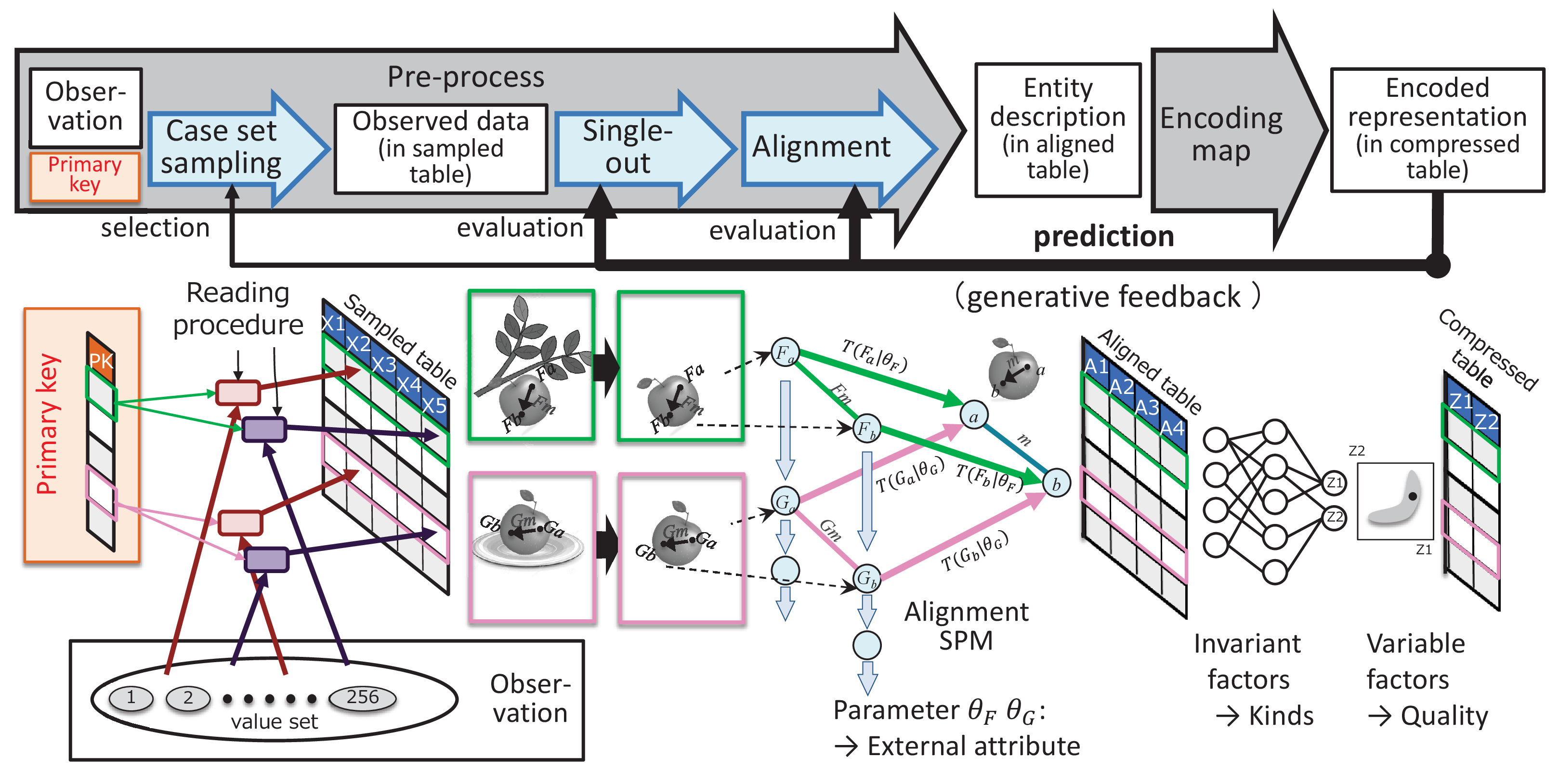} 
\caption{Mechanism of Entification: Mainly consists of a pre-process and an encoding map.}
\label{fig3}
\end{figure*}

The entification mechanism presented in this section consists of an encoding map and preprocesses to obtain an alignment table (Fig. \ref{fig3}).
The preprocesses include sequential processes corresponding to the functions of the case set sampling, singling out, and alignment (the order of these can be changed).
Any of the preprocesses are adjusted to increase the compression ratio in the following encoding.

\subsection{Case set sampling process} 

The case set sampling function, which can flexibly generate table format data, can be realized by the following mechanism proposed in \cite{Yamakawa2021-jp}.
As shown on the left side of fig.\ref{fig3}, this is "a mechanism consisting of a set of reading procedures that read values from a specific value set according to each value of the primary key.”
 
The sampled table obtained by this mechanism automatically satisfies the following three inductive inference premises \cite{Yamakawa2011-bx}.
Fist, {\bf Specifying relation}: A relationship in which fields on different columns in the same case can be mutually specified.
Second,  {\bf Equivalence of specifying-relation equivalence}: A second-order relation in which the same specifying relation holds even if the case is changed.
Third, {\bf Comparability}: A relation that enables at least value identity or similarity to be determined between any fields on the same column.

Because the sensor information in a particular modality satisfies the general properties of remote sensing, all reading procedures are read from a single value set, as shown in Fig. \ref{fig3}.
However, it is acceptable to refer to a different value set for each type of reading procedure in cases such as sampling a case consisting of multiple modalities.

\subsection{Single-out process}

Single-out function is realized by a process that
 removes the background of an individual in object recognition from images.
In general, the process extracts spatiotemporally adjacent regions contained within the boundaries of an individual.
Single-out processing can be divided into two main types: one is to segment the range of information from time and space for regions with locally similar features, such as image textures.
However, this approach renders it challenging to separate multiple overlapping individuals.
The other approach combines methods to extract a range of information that is consistent with the data generated top-down from the encoding map on a global basis.

\subsection{Alignment process} 

The alignment process realizes the alignment function, which corrects the observation of distal entities to obtain the entity description on the aligned table.

This process is a common alignment SPM-based transformation within the same sensor modality.

In the figure, the patterns of the target F and G are transformed by alignment SPM $T()$.
Here, $\theta_F$ and $\theta_G$ are the transformation parameters and correspond to the external attributes such as the position and orientation of the target, as observed from the subject.

\subsection{Encoding map} 

The encoding map projects the data distribution range in the aligned table to the encoded representation.
The encoding map compresses information by extracting invariant factors as a kind from the repeated pattern of entity descriptions.
The remaining variable elements in the encoding representation are mapped to several qualities inheres in substantial particular.
The kind-specific sameness function is realized by the encoding map from a specific entity description to a certain site of the encoded representation.
The kind-classification function determines whether the site to which a particular entity description is mapped to the encoded representation is within the coverage of a particular kind.

As discussed in section 2, intelligent subjects model the world by compressing based on the redundancy of observed information.
Therefore, recognition and learning in all previous processes are optimized to increase the compression ratio of the encoding map.
A wide variety of methods are used for this optimization, including error backpropagation, maximum likelihood estimation, and Bayesian estimation.
For example, in predictive coding, optimization is performed to ensure that the predictions produced by the encoding map match the observed values.

\subsection{Entification and each sub-process}


Thus, each mechanism in pre-process is primarily responsible for each function: the case set sampling process is responsible for the case set sampling function for the flexible acquisition of tabular data, the alignment process is responsible for the alignment function to correct observations of distal entities, and the single-out process is responsible for the spatiotemporal localization function, which is a condition for diachronic identity.
The encoding map is responsible for the kind-classification and criterion of sameness, which are conditions for diachronic identity.


Tab. \ref{table1} shows the comparison of characteristics of the partial process, including context (e.g., task) dependency, modality dependency, specialized for remote sensors, hierarchy, spatiotemporal locality, usage for particular entification, and usage for universal entification.
As can be observed from the table, the encoding map is a generic process; however, each preprocess has different characteristics.

Considering this mechanism as a whole, preprocess and encoding map have different properties as follows:
preprocess can preserve patterns obtained from the outside world, although they cannot provide an evaluation for optimization.
By contrast, an encoding map can provide an optimization index by the compression ratio, although it loses patterns obtained from the outside world.
Therefore, preprocess, and encoding map are combined and complement each other to realize entification.

This entification mechanism can be repeatedly stacked and corresponds to the hierarchical structure in neocortical  circuits, transformer-type deep neural circuits, and others.

\begin{table*}[t]
    \centering
    \begin{small}
\begin{tabular}{|p{20mm}||p{15mm}|p{15mm}|p{15mm}|p{25mm}|p{20mm}|p{18mm}|p{15mm}|}
    \hline \hline
        Name of process & Context-dependency & Uni-/Multi- modal & Specific to sensor & Hierarchy & Spatio–temporal locality   & Particular & Universal   \\ 
        \hline \hline
        Case set sampling & Exist & Uni/Multi & No & Category hierarchies (is-a relation) & Single entity size  & isochronous/ synchronous sampling & set of particulars in one kind \\   
        \hline
        Single-out  & Somewhat & Uni & Yes & - & Local  & Execute  & -  \\ 
        \hline
        Alignment  & None & Uni & Yes & Compositionality (part-of relation) & Single entity size  & Execute  & -  \\
        \hline
        \hline
        Encoding map & Very little & Uni/Multi & Somewhat & Various hierarchy & Do not care   & Execute & Execute   \\ 
         \hline
         \hline
    \end{tabular}
        \caption{Characteristics of each sub-process}
    \label{table1}
    \end{small}
\end{table*}

\section{ Current Trends and Future of Entification} 

The degree of achievement, success factors, remaining areas, and technical obstacles of entification are investigated on the ontological sextet as a comprehensive technical map.

\subsection{Achievement of current ML-AI}
In current ML-AI, recognizing the substantial particular is making progress.
For example, research on physical entities from visual information, has been progressing \footnote{https://iclr.cc/virtual/2022/workshop/4554}.
One of the results of this technology is the ability to decompose a still image into meaningful components and combine them based on linguistic instructions (e.g., "an astronaut riding hose in a photorealistic style") to create an image that is not experienced \cite{Ramesh2022-le}.

In the following, the technical elements that support the success of deep neural networks are described.

\subsubsection{Encoding map} 


As mentioned in the subsection "Request to compress the world," compressing and modeling the world is a central theme in intelligence research, mainly concerned with encoding maps.
For example, autoencoder, which was recognized as a neural network trained by error backpropagation in the 20th century, was a network for compression.
Currently, a generative model \cite{Jebara2012-zj} is being developed that infers the encoded representation and predicts the observed information from it. The predictive encoding algorithm \cite{Rao1999-ym} is well-established in more neuroscientific areas.
Many related research topics exists, including self-supervised learning, the free energy principle \cite{Friston2006-ry}, the good regulator \cite{Conant1970-mh}, and emulation theory of representation \cite{Grush2004-hd}.

\subsubsection{Pre-process in general} 
All three sub-processes of the pre-process can be realized as an attention mechanism.
The attention mechanism was developed from neocognitron and matured as a pooling layer in the convolutional neural network. 
In another stream, a sequential attention mechanism was proposed for multiple objects in a still image \cite{Eslami2016-ke}.
Furthermore, since 2017, deep neural network have been actively incorporating attention mechanisms \cite{Vaswani2017-dm}.
Today, transformer architectures that make extensive use of attention mechanisms are successful.

\subsubsection{Alignment method} 
Contrastive learning \cite{Chen2020-gi}, which was developed as a self-supervised learning for images, learns to move representations derived from individuals with augmented images closer together and images derived from different individuals farther apart to acquire an Entity description. 
This technique can learn alignment SPM for visual processing.
Advances in research on recognizing the position of each part of an artifact with moving parts \cite{Kawana2022-xc} suggest that hierarchical alignment processing of part-of-relationships (part-of-relationships) is also being realized.

\subsubsection{Single-out method} 
Semantic segmentation, which crops out localized regions with similar image features, is progressing. Contrastive learning, which provides identity information through data expansion, maps pixel-by-pixel cropped image regions to text \cite{Li2022-hh}. Semantic pixel embeddings can be generated to create object masks \cite{Van_Gansbeke2021-bo}.
Instance segmentation has been studied as a technique for identifying single-out objects. For example, when counting the number of chairs in a room, if the observed object is in the range corresponding to a chair in the encoded representation, it can be added to the count as a chair.
Furthermore, panoptic segmentation can separate objects and stuff (ex. ground, sky, sea, etc.) in a still image and assign IDs only to objects \cite{Kirillov2019-ui}.

\subsubsection{Case set sampling method} 
In entification for a particular entity, the case set is often sampled isochronously, depending on the characteristics of the sensor.
In spoken language processing, for example, synchronous sampling is performed on phonemes or words.
In any case, sampling in these cases is determined bottom-up by the designed algorithm and thus does not pose the challenge of finding new case set sampling.

\smallskip
Therefore, all the processes that support entification of substantial particular are being realized by ML-AI.
In actuality, the preprocess and encoding map correspond to the attention and full connect layers of the transformer.

\subsection{Future entification development along the sextet}

Considering these developments, ML-AI will be able to handle other categories of ontological sextets in the future.

\subsubsection{Advance to Substantial universal }

As mentioned previously, the kind, substantial universal is a specific range on the encoded representation. Therefore, its foundation has already been realized.

To date, however, we have not been able to properly handle hierarchies based on inheritance relations (is-a relations) between kinds. For example, cats and dogs belong to mammals, and mammals and fish belong to vertebrates.
Here, the kinds hierarchy essentially depends on the coverage of the attributes of the substantial universal. 
In other words, concrete entities have a rich set of attributes, whereas abstract entities cover a wide range of concepts.
For example, whereas the attribute "oviparity" is common to all mammals, howling is an attribute unique to dogs.

Studies are already underway to classify substantial universal by providing teacher labels for kinds. For example, some studies distinguish between kinds of vehicles and kinds of animals\citep{Van_Gansbeke2021-bo}.
In addition, the situation decomposition method can extract a hierarchy of kinds through unsupervised learning \cite{Yamakawa98:0}.
Based on these previous studies, research will proceed on encoding maps that can acquire a hierarchy of kinds using unsupervised learning.

\subsubsection{Advances to Quality: Particular and Universal }

Quality particular is an entity that depends on a substantial particular, e.g., this headache, this tan, etc.
Each quality value can be regarded as part of the variables that appear in the encoded representation. 
In the late 2010, ML-AI to extract quality was being studied as a research topic called "disentanglement."
The typical task was moving MNIST as well as face recognition and generation.
However, at that time, the single-out method was not mature; therefore, independent components were often extracted from the entire video scene.
This sometimes led to confusion between external attributes, which are parameters in the alignment process (e.g., orientation and speed in a figure image), and internal attributes (quality), which were inherent in the substantial particular.
In the future, the entification of quality will progress on the scaffold of the entification of substantial particular.

Next, quality universal is an entity that differentiates substantial universal.
For example, in the quality of "blueness," the blueness of the sky and the blueness of cars are common.
To build a mechanism for recognizing the quality universal in the future, it is necessary to have a mechanism to acknowledge quality beyond the kind of substantial entities.

\subsubsection{Advances to Occurrent (Process): Particular and Universal}

Occurrent(process) on ontology sextet is a time-evolving entity at some time interval (temporal boundary), often associated with causality.
Process universal, for example, collision, walking, mating, transmission, movement, influx, and support, etc., convey information about processes particular, such as this collision or this walk.

Because occurrent has a time interval, its entification has the following two difficulties compared to continuant entification.
The first is the issue of determining the time width of the cases to be retrieved in the case set sampling.  For example, to capture a scene of a ball hitting a wall, it is necessary to determine the timing of the beginning and end points in each case. Furthermore, it is necessary to define a primary key to identify every case in the set.
The second is the amount of information handled in each case.
Even if it is occurrent, once the time-varying information is deployed in memory, various computational processes can be performed on it as a static object, just as in the case of a continuum.
However, by these deployments, the memory capacity and processing to handle it will be enormous, so in reality, it is necessary to perform appropriate abstraction.
The commonsense policy suggested by ontological sextet is to describe a process particular in a way that multiple substantial particulars participate in it. 
In the example of a ball colliding with a wall, the amount of processing can be reduced by focusing only on the external attributes (position, speed, etc.) related to the participating wall and ball.
There are other issues to be addressed in handling occurrent, although it is expected to develop rapidly as it has been activated in recent years\footnote{https://crl-uai-2022.github.io/}.

\section{Conclusions} 
This paper proposes a set of goals to enable AI to comprehensively recognize all categories on the ontological sextet as an indicator to evaluate AI's reaching the human level (comprehensive technology map approach based on ontology).
Entification, the recognition of existence, is the process of finding patterns (often in the distance) preserved in the world.
Furthermore, we identified the necessary functions for entification to recognize the existence of a substantial particular, including the requirement to compress the information of the world, the condition of diachronic identity, the function to obtain observation data in a table format, and the function to correct the observation of a distal existence.
We then proposed a mechanism for preprocessing, consisting of the case set sampling, single-out, alignment sub-processing, and an entification mechanism, consisting of an encoding map, to satisfy these necessary functions.
Recent ML-AIs (e.g., deep learners with transformer-type architectures) can recognize and further manipulate the substantial particulars because such AIs cover all the aforementioned subprocesses and integrate them.
In the future, there are three main directions to expand the coverage of entification on the  sextet.
Substantial universal is already expressed as an invariant factor of the encoding map, although expanding it to handle concept hierarchies based on inheritance relations will be necessary.
Quality universal has already been expressed as a variable factor in the encoding map. However, it will be necessary to develop a mechanism to extract universal attributes (e.g., color, shape, weight, etc.) beyond kinds.
In occurrent (process), it is necessary to handle the dynamic behavior of the participating "partial particulars." Some issues are still to be addressed. However, enabling the entification of occurrent is related to understanding causal relationships worldwide and will be a significant foundation for intelligence to solve various tasks creatively.

When each category of the ontological sextet is further subdivided, it may be necessary to consider concepts that have not been handled in the previous discussions.
Some  well-established upper-ontologies include BFO \cite{Smith2005-mz}, DOLCE \cite{Gangemi2002-ko}, YAMATO \cite{Mizoguchi2022-qy}, and SOWA \cite{Sowa1995-sy}. A special issue was conducted recently \cite{Borgo2022-wm}.
Some difficulties exist in recognizing the existence of boundaries, domains, functions, and roles in time and space contained in these upper ontologies.
However, because there are not many categories in the hierarchy of upper ontologies, the goal of making all of them recognizable will remain valid.
This is because differences in the treatment of entification by ML-AIs occur when the higher ontological categories are different, rather than by detailed kinds, such as dogs and cats.

In the future, AI is expected to recognize every entity. The major remaining obstacle is the entification of occurrent, although no critical difficulty can be observed. 
Therefore, the time required to overcome it will be approximatley five more years. The entification of advanced concepts such as space and function will progress in parallel. Thus, human-level AI is expected to emerge by 2030, capable of recognizing nearly any kind of existence.

Based on this study, the following are the several directions for the future development of deep learning.
First, compared to the focus on considering variable factors obtained from the encoding map as attributes, less attention is paid to invariant factors. However, in the case set sampling, the potential value of invariant factors as kinds is relatively large. Humans have developed the science by discovering new kinds, such as cells, elements, and black holes.
Second, the three sub-processes of the pre-process are mixed in the transformer as the attention layer because they are all information selection processes. However, they are very different in terms of context, modality, hierarchy, and other properties. Therefore, conducting a study that distinguishes between the subprocesses may be helpful.
Third, case set sampling for recognizing specific objects from visual information is pre-designed; thus, there is no need to examine these functions in depth. However, in a task such as counting the types of playing cards (there are four different answers), it is necessary to control the sampling of each kind of card. Therefore, a flexible case set sample still has enormous potential for further research.
Fourth, transformer's successful visual object recognition and spoken language recognition assume that objects and speakers are preserved to some degree at the distal level during recognition. By contrast, in natural language processing, which transformer first succeeded in, the nature of what is preserved at the distal level is unknown. One possibility might be the speaker's personality. Therefore, it might be helpful to use character consistency to perform contrastive learning.
Fifth, setting different identity criteria for a different kind may be helpful, e.g., in object recognition techniques.


\newpage

\bibliography{aaai23.bib}

\end{document}